\documentclass{IEEEoj-data}
\usepackage{cite}
\usepackage{multicol}
\usepackage{multirow}
\usepackage{hyperref}
\usepackage{amsmath,amssymb,amsfonts}
\usepackage{algorithmic}
\usepackage{graphicx,color}
\usepackage{textcomp}
\def\BibTeX{{\rm B\kern-.05em{\sc i\kern-.025em b}\kern-.08em
    T\kern-.1667em\lower.7ex\hbox{E}\kern-.125emX}}
\AtBeginDocument{\definecolor{ojcolor}{cmyk}{0.93,0.59,0.15,0.02}}

\begin{document}
\receiveddate{00 XXXX, 20XX}
\reviseddate{00 XXXX, 20XX}
\accepteddate{00 XXXX, 20XX}
\publisheddate{00 XXXX, 20XX}
\currentdate{00 XXXX, 20XX}
\doiinfo{DD.20XX.XX.XX}

\title{DataDRILL: Formation Pressure Prediction and Kick Detection for Drilling Rigs}

\author{Murshedul Arifeen\authorrefmark{1,2}, Andrei Petrovski\authorrefmark{1,2,3}, Md Junayed Hasan\authorrefmark{1,2,4},Igor Kotenko\authorrefmark{3,5}, Maksim Sletov\authorrefmark{3}, Phil Hassard\authorrefmark{7}}
\affil{National Subsea Centre, Aberdeen, AB21 0BH, Scotland, United Kingdom}
\affil{School of Computing, Engineering, and Technology, Robert Gordon University, Aberdeen
AB10 7AQ, Scotland, UK}
\affil{Faculty of Secure Information Technologies, ITMO University, St. Petersburg, 197101, Russia}
\affil{School of Social \& Env. Sustainability, University of Glasgow, Dumfries DG1 4ZL,  Scotland, United Kingdom}
\affil{Federal Research Center of the Russian Academy of Sciences (SPC RAS), St. Petersburg, Russia}
\affil{Energy Transition Institute, Robert Gordon University, Aberdeen, AB10 7GJ, Scotland, United Kingdom}

\corresp{CORRESPONDING AUTHOR: Murshedul Arifeen (e-mail: m.arifeen@rgu.ac.uk)}
\authornote{The authors contributed equally to this article.}
\markboth{Preparation of Papers for IEEE DATA DESCRIPTIONS}{Author \textit{et al.}}

\begin{abstract}
Accurate real-time prediction of formation pressure and kick detection is crucial for drilling operations, as it can significantly improve decision-making and the cost-effectiveness of the process. Data-driven models have gained popularity for automating drilling operations by predicting formation pressure and detecting kicks. However, the current literature does not make supporting datasets publicly available to advance research in the field of drilling rigs, thus impeding technological progress in this domain. This paper introduces two new datasets to support researchers in developing intelligent algorithms to enhance oil/gas well drilling research. The datasets include data samples for formation pressure prediction and kick detection with 28 drilling variables and more than 2000 data samples. Principal component regression is employed to forecast formation pressure, while principal component analysis is utilized to identify kicks for the dataset's technical validation. Notably, the R2 and Residual Predictive Deviation scores for principal component regression are 0.78 and 0.922, respectively.\\ 
 \\ 
 {\textcolor{ieeedata}{\abstractheadfont\bfseries{IEEE SOCIETY/COUNCIL}}}     Industry Applications Society\\  
 \\
 {\textcolor{ieeedata}{\abstractheadfont\bfseries{DATA DOI/PID}}}   10.5281/zenodo.12759014\\ 
  
 {\textcolor{ieeedata}{\abstractheadfont\bfseries{DATA TYPE/LOCATION}}}   OTR simulator, School of Computing, Engineering, and Technology, Robert Gordon University, Aberdeen, UK

\end{abstract}


\maketitle

\section{BACKGROUND}

\begin{figure*}[htbp]
    \centering
    \includegraphics[width=0.50\linewidth]{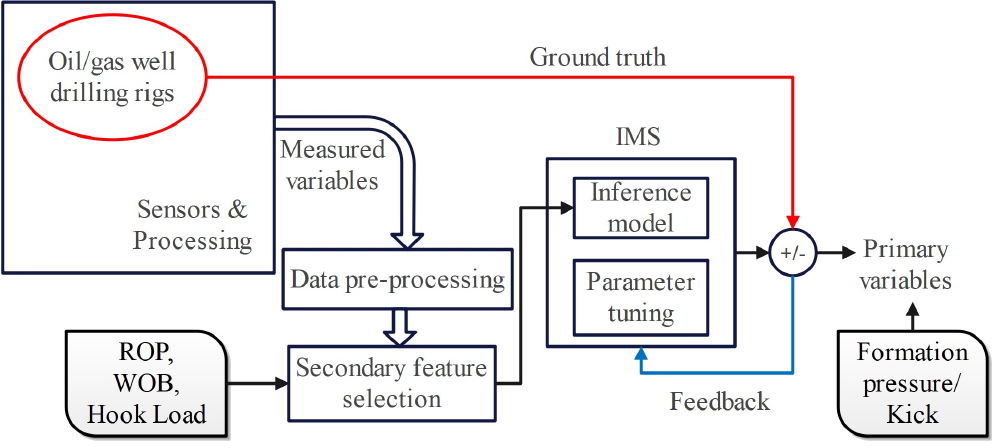}
    \caption{Overview of the inferential measurement system (IMS) architecture for drilling rigs. The sensors and actuators are connected to the drilling rigs to measure different drilling variables. After preprocessing, the secondary variables related to the primary variable are chosen. The inference model is then used to predict the primary variable, such as formation pressure from the secondary variables. The inference model can also take feedback from the predicted data and update its parameters. }
    \label{fig:IMS}
\end{figure*}
Petroleum engineering involves drilling wells to discover and extract hydrocarbons like crude oil and natural gas \cite{sircar2021application}. Engineers encounter different formations and pressures during drilling through specific geological columns, including pore/formation and fracture pressure \cite{abdelaal2021data}. Formation pressure, exerted by fluids in porous media, is the pressure within rock pore space \cite{abdelaal2022real, jingyi2023rock}. At a particular depth, the normal formation pressure gradient ($0.433$ psi/ft for freshwater to $0.465$ psi/ft for saltwater) is influenced by the weight of the saltwater column from the surface to the point of interest \cite{abdelaal2021data, abdelaal2022real, Korsah}. The normal pressure in underground formations is variable and is influenced by factors such as dissolved salts, fluid types, gas presence, and temperature gradient \cite{abdelaal2022real}. Any deviation from the usual pressure pattern can be subnormal or overpressure. When the pressure in a formation exceeds the hydrostatic pressure, it is termed supernormal or overpressure \cite{abdelaal2022real}. Supernormal pressure results from normal pressure and an additional pressure source (e.g., geological, mechanical, geochemical, geothermal, and combined reasons), while subnormal pressure occurs when the pressure is lower than normal \cite{abdelaal2021data, abdelaal2022real, zhong2020prediction}. Supernormal or overpressure may result in kicks, blowouts or unexpected influx, while subnormal pressure may cause differential pipe sticking or circulation loss \cite{abdelaal2021data, li2012pore}. Therefore, understanding subsurface formation pressure variations is critical for refining well trajectory, crafting precise drilling plans, and assessing wellbore stability for oil and gas wellbores \cite{abdelaal2021data, abdelaal2022real, li2012pore}. Moreover, accurate formation pressure estimation improves drilling operations, prevents hazards such as circulation loss and kicks, and reduces drilling time and costs \cite{farsi2021predicting}.

Formation pressure estimation can be achieved through either empirical or data-driven models based on drilling variables, well-log data, or formation characteristics, which fall into the inferential measurement or soft/virtual sensors-based systems \cite{ahmed2019new}. Empirical or mathematical models are challenging to develop and lack dynamism, whereas data-driven models leveraging artificial intelligence (AI) or deep learning are considered more robust and efficient. Inferential measurement systems (IMS) based on data-driven models process data from physical sensors and infer more complex system characteristics, such as the maintenance-free operation period \cite{souza2016review}. Figure \ref{fig:IMS} shows the working mechanism for IMS system. After gathering the data, a set of secondary variables is selected (through sampling, normalisation, noise reduction, and feature selection) and used to construct inferential models, effectively functioning as virtual sensors. These models enable users to estimate the primary variable or more complex characteristics that are not directly measurable, such as the formation or pore pressure in the well. Artificial neural networks based soft sensors, particularly feed-forward neural networks, have recently gained popularity among the research community in predicting formation or pore pressure from the correlated drilling variables (e.g., Rate of penetration (ROP), Weight on bit (WOB), Hook load, torque, etc.) or well-log data \cite{rashidi2018artificial, keshavarzi2013real, hu2013new, khaled2022new, ahmed2019new, kianoush2023ann}. However, among these studies, several authors only used a small number of data samples with few variables to train and test the neural network-based regression models. A small dataset can cause overfitting in a neural network model, leading to poor performance with new testing data and low generalisability \cite{safonova2023ten}. Although neural networks are commonly utilised for pore pressure prediction, only a few researchers have conducted experiments using classical machine learning (ML) models such as support vector machines, random forests, quantile, ridge, and XGBoost \cite{booncharoen2021pore, yu2020machine, matinkia2022novel, ahmed2019comparative}. Nevertheless, neural network-based models excel in learning complex patterns and demonstrate better generalisation ability than these classical ML models. Neural networks are not only utilised in predicting formation pressure but are also widely accepted for detecting kicks in the wellbore. Recently, several authors have used various types of neural network-based models for detecting kicks through IMS, such as physics-informed neural networks \cite{sha2024automatic}, parameter adaptive neural networks \cite{zhang2024intelligent}, convolutional neural networks \cite{chen2024kick}, and recurrent neural networks \cite{wang2023time}. 

However, the inferential measurement research field for offshore drilling rigs has not yet advanced as much as other domains, such as the chemical and process industries, where advanced AI algorithms are used to predict hard-to-measure primary variables \cite{souza2016review}. The dataset is crucial in developing, training, and validating advanced AI models for enhancing well-drilling research. Existing literature on data-driven inferential models for drilling rigs has only utilised small datasets to investigate the primary variable prediction problem, which may lead to model overfitting. Additionally, most datasets used in these research studies are not publicly accessible due to confidentiality agreements. The unavailability of a publicly accessible dataset could impede the advancement of automation research in drilling rigs. A publicly accessible dataset is essential in evaluating the efficacy of current methodologies, facilitating technological progress, and enriching educational initiatives within this domain. Public datasets can set the standards for evaluating IMS models used in drilling rig research. They can also help identify and assess new models for automating drilling rigs and provide valuable educational resources for researchers and students to understand drilling rig complexities.

This paper presents two datasets of drilling rigs for predicting formation pressure and detecting kicks. To the best of our knowledge, this is the first public dataset for research on AI-enabled models of offshore drilling rigs generated from a digital twin of a drilling rig. It comprises $28$ drilling variables and more than $2000$ data samples. This dataset can significantly contribute to the research community by facilitating the development, training, and testing of AI models for predicting formation pressure and detecting kicks. To validate the technical aspect of the dataset, we have utilised principal component analysis (PCA)-based models to predict formation pressure and detect kicks.

\section{METHODS AND DESIGN}
In the following section, we have outlined the well-drilling process, the experimental setup for data collection, and two engineering scenarios for dataset generation. Scenario 1 explains the formation characteristics for formation pressure prediction, while scenario 2 delineates the formation characteristics for kick detection.
\subsection*{Well Drilling}
\begin{figure*}[htbp]
\centering
\includegraphics[width=0.50\linewidth]{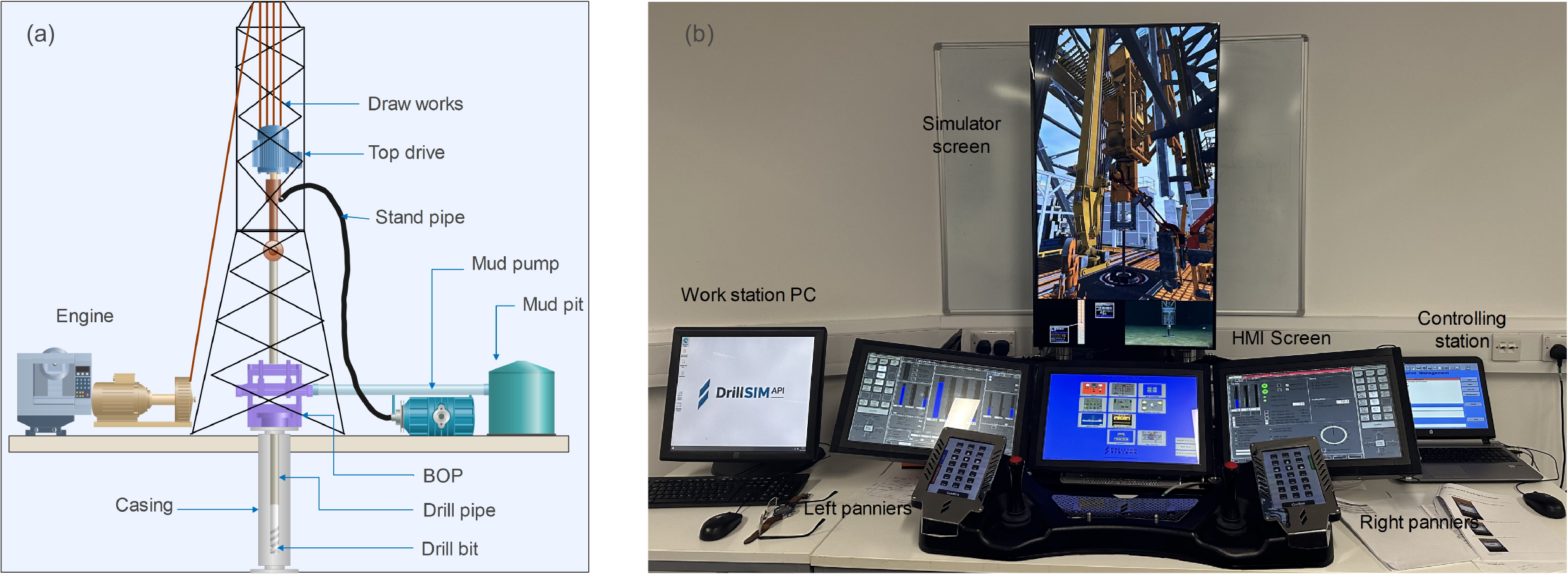}
\caption{OTR simulator in the experimental room. The main simulation computer is connected to a workstation PC, HMI, and an instructor laptop. The workstation PC hosts the API for different rig packages and is used to collect data from the simulator. The instructor's laptop is used to set drilling scenarios based on geological data.}
\label{fig:OTRSim}
\end{figure*}

Offshore oil well drilling rigs operate by accessing and extracting oil reserves beneath the ocean floor. In this subsection, the drilling process \cite{PONTEJR202155} is briefly explained. The process begins with positioning the rig, which can be either a floating structure or a fixed platform anchored to the seabed. Drilling commences using a drill bit attached to a long string of drill pipes hooked from a draw-works, penetrating the ocean bed through rotational and hydraulic force from the top drive. Drilling mud, a specially formulated fluid, is continuously pumped through the drill pipes to cool the drill bit, lubricate the passage, and bring rock cuttings to the surface. As the well drills deeper, casing pipes are inserted to prevent collapse and isolate underground pressure zones. Advanced technologies such as dynamic positioning systems and blowout preventers maintain the rig's stability and ensure safety by preventing uncontrolled oil or gas eruptions. The entire operation is monitored and controlled from the rig's control room, ensuring efficient and safe oil extraction from beneath the ocean floor. Figure \ref{fig:OTRSim}(a) shows a schematic diagram of a basic drilling rig with different components.

We conducted experiments to generate datasets for formation pressure prediction and kick detection in the On the Rig (OTR) simulator \cite{OTRSimWebPage}. The OTR from 3t Global Drilling Systems is a real-time portable simulator replicating drilling and equipment operations, well control, and crane training on various rigs, including Land, Jackup, DrillShip, and Semi-submersible rigs, for research, experimental and training purposes. We have used the DrillShip module for our experimental setup. The OTR simulator PC is linked to a workstation PC, simulator screen, human-machine interface (HMI), and a controlling station/laptop. Figure \ref{fig:OTRSim}(b) shows the OTR simulator in the experimental room at Robert Gordon University. The workstation PC hosts an application programming interface (API) with multiple rig control packages for conducting well-control research and experiments. Each package contains various variables related to the drilling rig and downhole infrastructures. The controlling laptop is used to configure a specific drilling scenario and initiate a drilling exercise in the simulator PC. For example, rock/formation parameters were configured on the controlling laptop for formation pressure prediction and kick detection problems. The primary purpose of the dataset generated from the OTR is to provide evidential support to the drilling operators on the consequences of their actions, which are currently taking place offline \cite{Arifeen:InsiderThreat:2024}. However, it is possible to hybridise the available OTR-based digital twin with physical data acquisition devices via networks of PLCs and engineering workstations.  The experience of providing synchronisation facilities with digital twins comes from building and using cyber ranges in research and teaching at ITMO University (St. Petersburg, Russia) \cite{CyberRange}.

\subsection*{Scenario1 (Formation Pressure Prediction)}
In our experiment to generate a comprehensive dataset for predicting formation pressure, we considered a 4-foot (ft) formation/rock with five distinct formation types. The characteristics of these formation types were defined in the controlling station. The designed formation/rock was precisely located at $12679$ ft, aligning the drilling bit at the same depth. The drilling bit was rotating by a top drive of $80$ rpm. Below $12679$ ft, the formation type was the Forties, adding a layer of complexity to our dataset. The Forties Sandstone comprises fine to coarse sandstone interlayered with medium to dark grey siltstone and mudstone \cite{FormationLayer}. The sandstone is moderately to poorly sorted and sometimes pebbly. The Forties formation exerts 6050 psi pressure at a depth of $12679$ ft and has a porosity pressure of 0.55 psi/ft. The permeability of the Forties layer is set to 1 millidarcy (md) with water as the fluid type. The abrasion factor and drill value are set to $0.3$.

\begin{figure*}[htbp]
\centering
\includegraphics[width=0.50\linewidth]{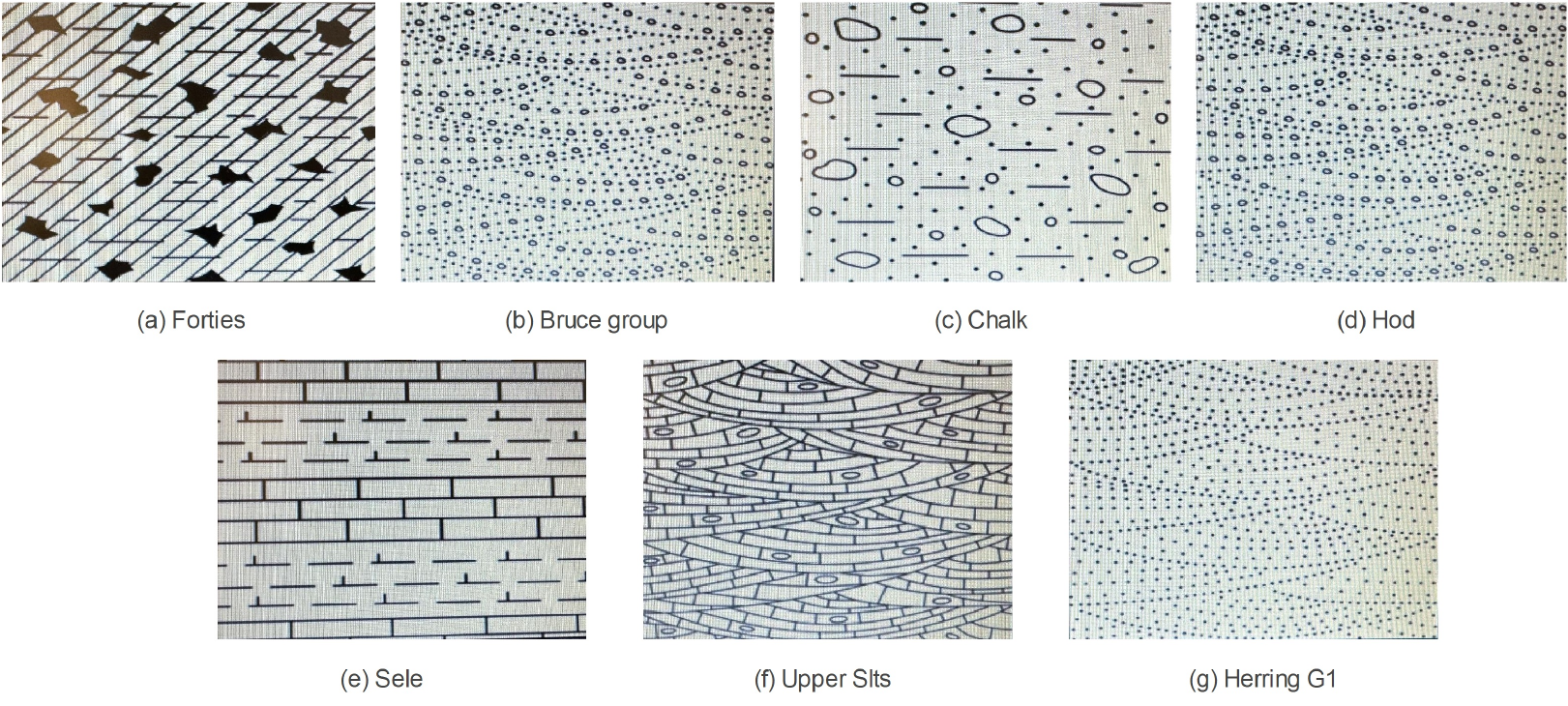}
\caption{Patterns of different layers of the designed formation for Scenarios 1 and 2 (Figures are taken from the simulator)}
\label{fig:Layers_pattern}
\end{figure*}

The next layer in the formation is set to Bruce Group and possesses a rock strength of $0.3$. This layer exerts a pressure of $7603$ psi with a porosity pressure of $0.60$ psi/ft, significantly higher than the preceding layer. The fluid type present in this layer is Gas, with a permeability of $50$ md, indicating that Gas in the Bruce Group flows much more rapidly than water in the Forties layer. The abrasion factor assigned to this layer is $1$, denoting the highly abrasive nature of the Bruce Group. The subsequent layers, Chalk, Hod, and Herring G1, have rock strength, abrasion factor, and drillability values set to $0.1$, respectively. The pressure exerted by Chalk and Hod is $6721$ psi and $4693$ psi, gradually decreasing from the Bruce group layer. The pressure then increases to $19052$ psi after $12684$ ft at the Herring G1 layer. For these last three layers, the fluid type is Gas with a permeability value of $1$ md. Therefore, the Gas in these layers flow slowly than the Bruce group layer. Table \ref{tab:FormationPressure} shows the summary of the values set for different layers of the designed formation in the controlling laptop. Figure \ref{fig:Layers_pattern} illustrates the patterns of the different formation layers used in this experiment.

% Table generated by Excel2LaTeX from sheet 'Sheet3'
\begin{table*}[htbp]
  \centering
  \caption{Formation characteristics or parameters for Scenario 1}
    \begin{tabular}{ccccccccccc}
    \hline
    \textbf{LN}&\textbf{FType}&\textbf{FD (ft)} & \textbf{MD (ft)} & \textbf{Drill} & \textbf{AF} & \textbf{Fluids} & \textbf{Perm (md)}  & \textbf{PP (psi/ft)} & \textbf{Pressure (psi)} & \textbf{RS} \\
    \hline
    \hline
    1&Forties    &$11000$  & 11000  & 0.3   & 0.3   & Water & 1.00  & 0.55  & 6050   & 2.0 \\
    \hline
    2&Bruce Group&$12680$ & 12680  & 0.3   & 1.0     & Gas   & 50.00 & 0.60  & 7603   & 0.3 \\
    \hline
    3&Chalk       &$12682$&12682 & 0.1   & 0.1   & Gas   & 1.00  & 0.53  & 6721   & 0.1 \\
    \hline
    4&Hod          &$12684$ & 12684 & 0.1   & 0.1   & Gas   & 1.00  & 0.37  & 4693   & 0.1 \\
    \hline
    5&Herring G1   &$30003$ & 30003 & 0.1   & 0.1   & Gas   & 1.00 & 0.64 & 19052   & 0.1 \\
    \hline
    \hline
    \end{tabular}%
  \label{tab:FormationPressure}%
  \vspace{0.3cm}

  Legend: LN-- Layers number, FType--Formation Type, FD-- Formation Depth, MD--Measured Depth, Drill-Drillability, AF--Abrasion Factor, Fluids--Fluid Types, Perm--Permeability, PP--Porosity Pressure, RS--Rock Strength
\end{table*}%

\subsection*{Scenario 2 (Kick Detection)}
We have developed a $10$-foot formation for a kick-detection scenario at the controlling station. Table \ref{tab:KickDetection} summarises the formation's characteristics. The formation begins at a depth of $12641$ ft with the bit at the same position. For this scenario, we set the top drive speed to $110$ rpm. Before commencing drilling, the formation type was Sele. The first layer of the formation, up to $12643$ ft, is the upper slts. This layer applies a pressure of $6827$ psi with a gradient of $0.54$ psi/ft. The fluid type of this layer, with a drillability factor of $0.5$, is identified as Water, and it has a permeability of $1.00$ md. The second layer is the Forties, with a slightly higher pressure of $7262$ psi and a gradient of $0.57$ psi/ft compared to the previous layer. Like the first layer, this layer's fluid type is Water, with a permeability of $1.00$ md. Despite having a higher drillability factor than the previous layer, this layer exhibits a low abrasion factor of $0.3$. Bruce group and chalk are selected for the subsequent two formation layers. The Bruce group applies a pressure of $8223$ psi at $12650$ ft with a pore pressure/ gradient of $0.65$ psi/ft. Both layers contain gas with a higher permeability of $50$ md than the previous layers.

As the drill bit reaches $12650$ feet within the Bruce group layer at a pressure of $8223$ psi, the well faces a significantly higher formation pressure than the hydrostatic pressure of $7006$ psi. This sudden and significant pressure disparity leads to a kick scenario in the well, a crucial indication of the entry of formation fluid into the wellbore. Table \ref{tab:DrillParaSc2} illustrates the changes in drilling parameters before and after the kick. Below 12650 ft, all three pumps (Pump $1$, $2$ and $3$) were pumping drilling fluids at $1749$ psi, $1740$ psi, and $1740$ psi, respectively. However, once the drill bit reached $12650$ ft, the pumping rates of all the pumps gradually increased to balance the formation and hydrostatic pressures. In contrast, based on the mud data (table \ref{tab:DrillParaSc2}), it is apparent that the return flow of the fluids also increased after the kick and reached $100\%$ over a very short distance of $12651.5$ ft. Moreover, the traditional indicator of kick (Pit gain-loss) has also significantly increased to indicate a kick.

% Table generated by Excel2LaTeX from sheet 'Sheet2'
\begin{table*}[htbp]
  \centering
  \caption{Formation characteristics or parameters for Scenario 2}
    \begin{tabular}{ccccccccccc}
    \hline
    \textbf{LN}&\textbf{FType}&\textbf{FD (ft)} & \textbf{MD (ft)} & \textbf{Drill} & \textbf{AF} & \textbf{Fluids} & \textbf{Perm (md)}  & \textbf{PP (psi/ft)} & \textbf{Pressure (psi)} & \textbf{RS} \\
    \hline
    \hline
    1&Sele &6185 & 6185 & 2.0     & 1.0     & Water & 10.0 & 0.53  & 3288    & 2 \\
    \hline
    2&Upper Slts &12641 & 12641 & 0.5   & 1.0    & Water & 1.0 & 0.54  & 6827  & 0.5 \\
    \hline
    3&Forties&12643 & 12642 & 0.8   & 0.3   & Water & 1.0 & 0.57 & 7262  & 0.8 \\
    \hline
    4&Bruce Group&12650 & 12648 & 0.5   & 1.0     & Gas   & 50.0 & 0.65 & 8223  & 0.5 \\
    \hline
    5&Chalk &12704 & 12704 & 0.5   & 1.0     & Gas   & 50.0 & 0.64 & 7622  & 0.5 \\
    \hline
    \hline
    \end{tabular}%
  \label{tab:KickDetection}%
  \vspace{0.3cm}

  Legend: LN-- Layers number, FType--Formation Type, FD-- Formation Depth, MD--Measured Depth, Drill-Drillability, AF--Abrasion Factor, Fluids--Fluid Types, Perm--Permeability, PP--Porosity Pressure, RS--Rock Strength
\end{table*}%

The drilling operation took $13$ to $15$ minutes to penetrate the planned rock/formation for both scenarios. Throughout the drilling process, we utilised the API from the workstation to collect time series data for various variables using OTR rig packages. The rig packages used for this experiment are \textit{WellControlManager}, \textit{FrictionLossInAnnulus}, \textit{SwabAndSurge}, and \textit{DownholeSloughing}. The package variables are associated with different functions based on the API definition. For example, the \textit{set()} method assigns an initial value to a variable for the simulation scenario. In contrast, the \textit{get()} method is employed to retrieve data from the running scenario in the simulator. After configuring the entire engineering setup from the instructor station, we've solely used the \textit{get()} method to retrieve the values of the chosen variables. 

The pseudocode outlining the algorithm defined in the API for generating the time series dataset is included in the supplementary document. We have defined the data-retrieval algorithm following the OTR rig package API standards. The algorithm is coded in $C\#$ within a console application of the API. This application contains a main class, \textit{WorkMain}, which encompasses two methods: \textit{Initialise()} and \textit{Update()}. The \textit{Initialise()} method is responsible for instantiating any rig package for simulation purposes. In contrast, the \textit{Update()} method enables us to define the variables of the instantiated packages and their corresponding functions for setting specific values or retrieving simulation data using the \textit{get()} function. Then, we created a custom function to generate a CSV file in a specified folder path and save the streaming time series data from the simulator.

% Table generated by Excel2LaTeX from sheet 'Sheet4'
\begin{table*}[htbp]
  \centering
  \caption{Drilling data for Scenario 2}
    \begin{tabular}{cccccc}
    \hline
          & \textbf{Variables} & \textbf{Below 12650 ft} & \textbf{At 12650 ft} & \textbf{At 12651 ft} & \textbf{At 12651.5 ft} \\
    \hline
    \hline
    \multirow{3}[14]{*}{\textbf{Pump data}} & Pump pressure (psi) & 1730 & 1731 & 1739& 1760 \\
\cline{2-6}          & Pump 1 speed (spm) & 99 &  99      &  99     &  99\\
\cline{2-6}          & Pump 1 pressure (psi) & 1749  & 1750 & 1758  & 1779 \\
\cline{2-6}          & Pump 2 speed (spm)& 99  &  99      &  99     & 99 \\
\cline{2-6}          & Pump 2 pressure (psi) & 1740  & 1741 & 1748 & 1769 \\
\cline{2-6}          & Pump 3 speed (spm)& 60  &   60    &     60  &  60\\
\cline{2-6}          & Pump 3 pressure (psi) & 1740 & 1741 & 1748 & 1769 \\
    \hline
    \multirow{2}[6]{*}{\textbf{Mud data}} & Return flow & \multicolumn{1}{c}{51\%} & 64\%  & \multicolumn{1}{c}{95\%} & \multicolumn{1}{c}{100\%} \\
\cline{2-6}          & Active Volume (bbl) & 198.9  & 198.9  & 200.7  & 204.9 \\
\cline{2-6}          & Pit gain loss (bbl)& -0.1 &       & 1.7 & 5.9  \\
    \hline
    \hline
    \end{tabular}%
  \label{tab:DrillParaSc2}%
\end{table*}%

\section{VALIDATION AND QUALITY}
\begin{figure}
    \centering
    \includegraphics[width=0.9\linewidth]{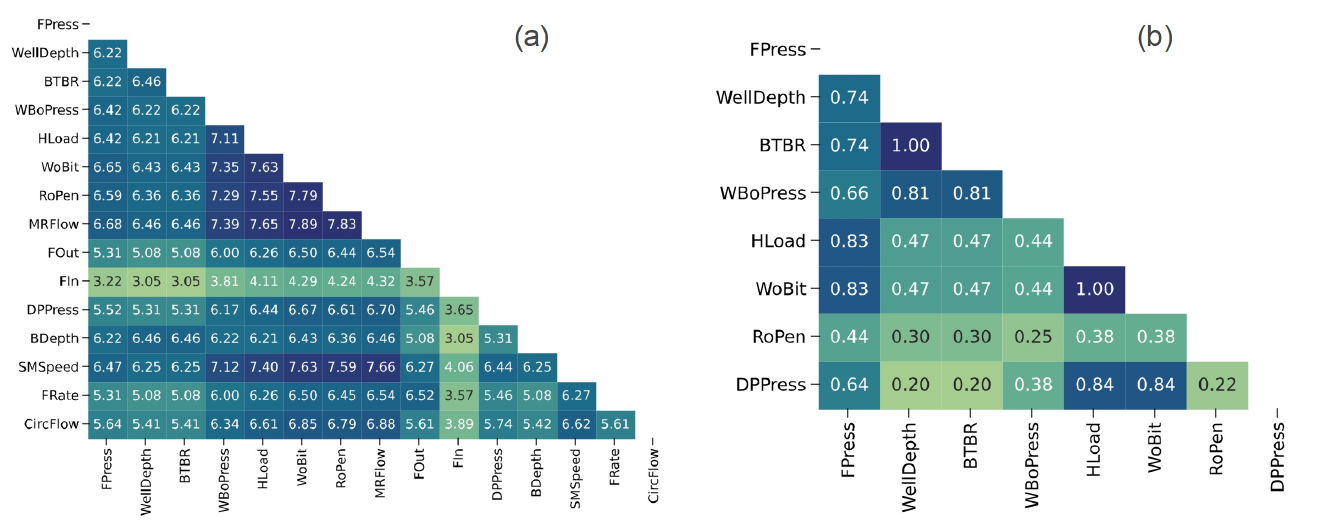}
    \caption{(a) Mutual information scores between the Formation pressure and other significant secondary variables. (b) Pearson correlation coefficient matrix between the Formation pressure and other highly correlated secondary variables}
    \label{fig:variable_relation}
\end{figure}

Soft sensor \cite{souza2016review} techniques can be helpful when predicting formation pressure. These models use secondary variables as input to forecast primary variables as output. In developing soft sensors, secondary variables are chosen based on their relationships with the primary variables. These relationships can be effectively measured using mutual information score (MIS), correlation coefficient, and neighbourhood distance techniques \cite{guo2020mutual, yuan2018deep}. Figure \ref{fig:variable_relation} visually represents these relationships regarding mutual information and Pearson correlation.

\begin{figure*}
    \centering
    \includegraphics[width=0.5\linewidth]{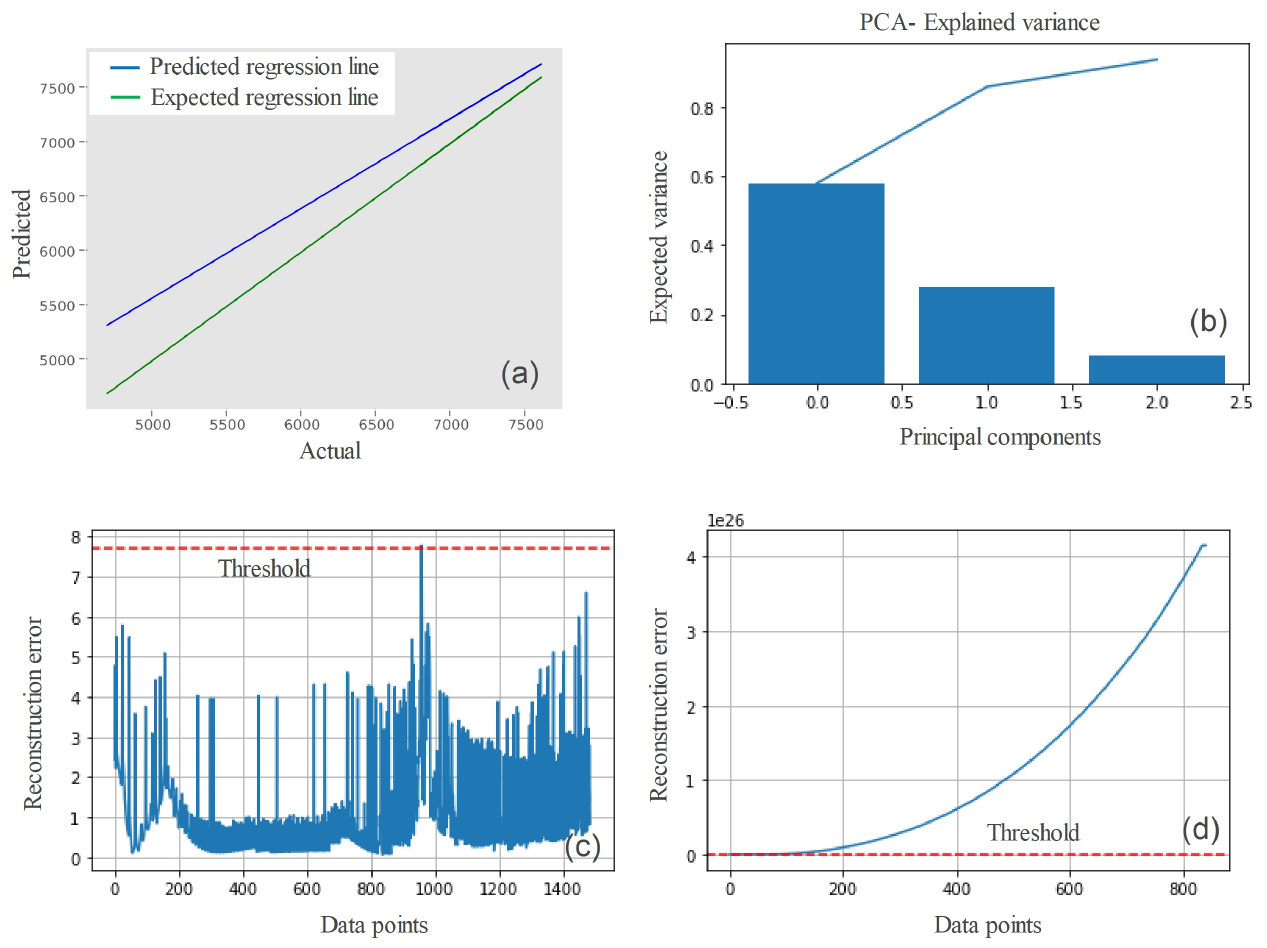}
    \caption{(a) Predicted vs. exprected regression line for principal component regression, (b) shows the expected variance for individual principal components and the cumulative variance, (c) reconstruction error for the training set, (d) reconstruction error for test set}
    \label{fig:kickPCA}
\end{figure*}

The correlation matrix is presented only for variables with a coefficient exceeding 0.4 with the variable Formation pressure. In contrast, the MIS matrix is displayed for variables with a mutual information score surpassing 3 with Formation pressure. According to the correlation matrix, Formation pressure is correlated with Well depth, BTBR, Wellbore pressure, Hook load, Weight on bit, Rate of penetration, and Drill pipe pressure. Notably, Hook load and Weight on bit exhibit a high correlation, with a value of 0.83. However, the MIS matrix indicates that the Formation pressure variable has significant mutual connections with several secondary variables, with the FIn attribute having the lowest score. We conducted a pilot experiment on principal component regression (PCR) to validate the formation pressure prediction problem using the highly correlated variables identified from the correlation coefficient matrix. Figure \ref{fig:kickPCA}(a) shows the expected and predicted regression lines for PCR regression model. Before training the PCR model, the secondary variables are made smooth through the savitzky golay filter. Then we fitted the PCR model with the secondary variables to predict the Formation pressure values. The R2 and RPD scores for the PCR regression model are $0.78$ and $0.9222$, respectively. However, more advanced methods, such as deep learning models, can be employed to improve prediction performance.

Kick-detection problems can be addressed in various ways, such as PCA, clustering methods, or deep learning models like Autoencoder. In this study, we conducted a preliminary experiment on PCA-based kick detection. First, the data samples before the kick occurs are separated from the dataset and defined as the training data. On the contrary, the data samples after the kick are considered test data. Also, the variables corresponding to the attributes such as CSDepth, BSize, FPress, CPress, MPS1, MPS2, MPS3, AMTD, and STP that do not change before and after the kick event are discarded from the dataset. Then, the PCA model is trained by reconstructing the original samples. The reconstruction error of the training data is then used to compute the threshold or kick detection limit. We chose the 99.99\% percentile for the training data to select the threshold. The threshold is used to determine the samples related to the kick event. Figure \ref{fig:kickPCA} (b) illustrates the cumulative expected variance ratio against the number of principal components. This figure demonstrates that two principal components can account for nearly 98\% of the variance. Figures \ref{fig:kickPCA} (c) and (d) also present the reconstruction error for the training and test data. From the figures, it is evident that the test data containing data samples after the kick incident are above the threshold, indicating that PCA can be a suitable option to detect kick in well drilling.

\section{RECORDS AND STORAGE}
The DataDrill dataset comprises two files representing the scenario 1 and 2 (available from \cite{Arifeen:DataDRILL:2024}). The formation and kick detection files are presented in CSV format and available for unrestricted access and download from this repository. The dataset for formation pressure prediction contains $2775$ records, while the kick detection file comprises $2338$ samples. Both datasets contain $28$ attributes or columns representing the variables of the OTR API. Table \ref{tab:DatasetVariables} shows the attributes with their descriptions. The attributes FRate and FDensity are extracted from the \textit{FrictionLossInAnnulus} package, while SMSpeed, BDepth, BSize, and MVis are acquired from the \textit{SwabAndSurge} Package. The CircFlow is taken from the \textit{DownholeSloughing} package and finally the remaining attributes of these datasets are chosen from the \textit{WellControlManager} package.

% Table generated by Excel2LaTeX from sheet 'Sheet1'
\begin{table*}[htbp]
  \centering
  \small
  \caption{Column names and brief description of the variables for both datasets}
    \begin{tabular}{cllp{0.5\linewidth}}
    
    \hline
    \textbf{Column no} & \textbf{Attribute names} & \textbf{Variables} & \textbf{Description} \\
    \hline
    \hline
    1     & CircFlow & Circulation Flow & The flow rate of the circulating fluid through the drill pipe  \\
    \hline
    2     & FRate & Flow Rate & The flow rate of the fluid from the mud pumps\\
    \hline
    3     & FDensity & Fluid Density & The density of the fluid circulating through the drill pipe \\
    \hline
    4     & SMSpeed &    String Moving Speed   &  The speed at which the string moves \\
    \hline
    5     & BDepth & Bit Depth &  Position of the drill bit from the surface \\
    \hline
    6     & BSize & Bit Size & The size of the drill bit \\ 
    \hline
    7     & MVis  & Mud Viscosity & The viscosity of the fluid flowing through the drill string \\
    \hline
    8     & FPress & Formation Pressure & The pressure exerted by the formation \\
    \hline
    9     & DPPress & Drill Pipe Pressure & Difference between the FPress at the bottom of the well and the hydrostatic pressure \\
    \hline
    10    & CPress & Casing Pressure & The pressure exerted by the surface on the casing inside the well \\
    \hline
    11    & FIn   & Flow In &   Amount of drilling fluid entering into the drill pipe \\
    \hline
    12    & FOut  & Flow Out & Amount of drilling fluid exiting the drill pipe \\
    \hline
    13    & MRFlow & Return Flow & The rate of the returned fluid flow \\
    \hline
    14    & ActiveGL & Active Gain Loss &  Indicator of the kick \\
    \hline
    15    & ATVolume & Active Tank Volume &  The volume of the mud tank for containing muds\\
    \hline
    16    & RoPen & Rate of Penetration &   Penetration rate of the drill bit \\
    \hline
    17    & WoBit & Weight on Bit & The amount of downward force exerted on the drill bit \\
    \hline
    18    & HLoad & Hook Load & The vertical force pulling down on the top-drive shaft at the bottom of the drill pipe \\
    \hline
    19    & WBoPress & Well Bore Pressure &   The pressure exerted to the formation from the well bore \\
    \hline
    20    & BTBR  &  Bit TVD below RKB &  The true vertical depth (TVD) of the bit below rotary kelly bushing \\
    
    \hline
    21    & MPS1  & Mud Pump Speed 1 & The speed at which the mud pump 1 is circulating the muds \\
    \hline
    22    & MPS2  & Mud Pump Speed 2 &  The speed at which the mud pump 2 is circulating the muds\\
    \hline
    23    & MPS3  & Mud Pump Speed 3 & The speed at which the mud pump 3 is circulating the muds\\
    \hline
    24    & AMTD  &   Active Mud Tank Density    & Density of the drilling mud of the tank \\
    \hline
    25    & ATMPV &   Active Tank Mud PV    &  The plastic viscosity (PV) of the drilling mud \\
    \hline
    26    & ATMYP &    Active Tank Mud YP   & Yield point (YP) of the drilling fluid inside tank \\
    \hline
    27    & STP   & Strokes Pumped & Measures the efficiency of the mud pumps  \\
    \hline
    28    & WellDepth & Well Depth & The height of the drilled well from the surface  \\
    \hline
    \hline
    \end{tabular}%
  \label{tab:DatasetVariables}%
\end{table*}%

\section*{INSIGHTS AND NOTES}
This dataset is free for academic, and research purposes. Users are allowed to copy, distribute, and transmit the dataset as well as to adapt and build upon it, provided that proper credit is given to the original creators. Proper attribution ensures that the creators receive credit (citations) for their work and encourages the ethical use of shared resources. Failure to comply with these citation requirements may result in restrictions on the use of this dataset.

\section{SOURCE CODE AND SCRIPTS}
Dataset and associated codes can be directly accessed from \cite{Arifeen:DataDRILL:2024}. 

\section{Acknowledgements}
This work is based on a study project initiated by a group of like-minded individual researchers and strategists from several institutes including Robert Gordon University, National Subsea Centre, the University of Glasgow, and ITMO. It is not a internally/externally funded project. The collective aim is to create future opportunities for researchers, particularly in the niche area of Oil and Gas rig digitization, and to explore potential markets.

\section{Declaration of generative AI and AI‑assisted technologies in the writing process}
During the preparation of this work, the author(s) used ChatGpt, Grammarly in order to improve the manuscript grammatically. After using this tool/service, the author(s) reviewed and edited the content as needed and take(s) full responsibility for the content of the publication.

\bibliography{main}
\bibliographystyle{ieeetr}

\end{document}